%% file: main.tex
\documentclass{article} %
\usepackage{iclr2027_conference,times}

\input{math_commands.tex}

\usepackage{amsmath}
\usepackage{amssymb}
\usepackage{bm}
\usepackage{enumitem}
\usepackage{booktabs}
\usepackage{multirow}
\usepackage{graphicx}
\usepackage{algorithm}
\usepackage{algorithmic}
\usepackage{tikz}
\usepackage{wrapfig}
\usepackage{hyperref}
\usepackage{url}

\usepackage{xcolor}

\title{Reconstructing LLM Lineage Trees \\ via Fine-Tuning Path Inference}

\author{Sota Ogawa \\ Institute of Science Tokyo \\ Tokyo, Japan \\ \texttt{ogawa.s.718d@m.isct.ac.jp} \And Shojiro Yamabe \\ Institute of Science Tokyo \\ Tokyo, Japan \And Daiki Nishiyama \\ Institute of Science Tokyo \\ Tokyo, Japan \And Jun Sakuma \\ Institute of Science Tokyo \\ RIKEN \\ Tokyo, Japan}

\begin{document}

\maketitle

\begin{abstract}
  Fine-tuned (FT) derivatives of large language models (LLMs) are now
  released in large numbers, yet their provenance relies on self-reports
  and cannot be verified. Misattributed lineage leaves a verifier unable
  to audit outsourced fine-tuning, confirm license compliance of derived models,
  or trace vulnerabilities that a parent passes to its children. All three require
  the direction of derivation, not merely its presence.
  Existing output-only methods are limited to pairwise derivation tests or
  undirected lineage estimation, while recovering a directed tree requires
  white-box access to model weights. We formalize the problem of
  reconstructing a directed model tree, including internal nodes, under
  gray-box access, where only teacher-forced log-likelihoods are observed
  and weights are never inspected. Fine-tuning raises the log-likelihood
  of the training data, and a child inherits this increase from its
  parent. Building on this property, we define an asymmetric per-task
  log-likelihood deviation between any two models as a trace of
  fine-tuning history, and derive from it an edge weight for each ordered
  pair of models. We propose TRACE-Edge, which exactly solves for the
  maximum spanning directed tree under these edge weights, with neither
  hyperparameters nor tree-structure constraints. We evaluate it on five
  configurations totaling 25 model trees built from three base models and
  seven tasks. TRACE-Edge achieves a parent-match accuracy of 0.74 when
  probed with the fine-tuning examples, and on average outperforms an
  output-similarity baseline even with held-out probes drawn from the
  same distribution as the training data.
\end{abstract}

\section{Introduction}\label{sec:intro}

Training a large language model (LLM) from scratch requires enormous
computational resources and training data. Fine-tuning (FT) a publicly
released pretrained model for a specific task is therefore widely used
instead. The spread of model-sharing platforms such as Hugging Face has
accelerated a development style that assumes models will be reused and
derived from one another.

When a model is fine-tuned to obtain another model, a parent-child
relationship holds between the two. Following these derivation
relationships as edges forms a tree structure rooted at a pretrained
model. Model-sharing platforms provide a mechanism for reporting each
model's source model, that is, its base model, through the model card,
but this report is optional and self-reported, and no mechanism exists
to verify its correctness.

Now that model derivation is commonplace, there are many situations
that require verifying a model's provenance. The first is auditing a
model after outsourcing fine-tuning to an external vendor. When a
company or research institution outsources FT to an external vendor,
the finished model is often accessible only through an API, and the
training process cannot be observed directly. In this situation, the
user has reason to verify whether the delivered model was in fact
derived from the specified base model, and whether updated models
supplied during operation continue to derive along the lineage
originally reported.

The second is tracking a model's license and rights. When a model is
derived from a base model that carries license conditions, confirming
compliance requires verifying the derivation relationship down to the
leaves of the tree structure, not just the direct child but its
grandchildren and further descendants. Exercising this right
effectively, however, presupposes the ability to verify which model a
given model was derived from. Here, showing that two models are
similar is not enough. Determining where the rights lie, or whether
they have been infringed, requires a method that establishes the order
in which the two models existed, that is, the direction of derivation.

The third is tracking the inheritance of a model's vulnerabilities and
safety risks. A parent model's vulnerabilities are inherited by its
child models through fine-tuning: adversarial prompts optimized on a
pretrained model transfer more effectively to its fine-tuned derivatives
than prompts optimized on unrelated models~\citep{tan2025leakawaypretrainedmodel}. Defenses against
derived models also depend on lineage information, since
\citet{zhong2026watch}'s backdoor detection needs the base model's
weights and degrades without them. This situation, too, requires not
just the presence of a derivation relationship but its direction.
Because fine-tuning is a one-directional operation that takes the
parent's parameters as its initial value, the propagation of
vulnerabilities is also asymmetric: a parent's vulnerability spreads to
its child, but not the reverse. An undirected derivation relationship
alone cannot pin down where the risk lies.

In situations like these, a verifier needs a way to verify a model's
provenance independently. A discrepancy between the reported lineage
and the true one can arise even without assuming any malicious intent
on the provider's part, for instance through an omitted or incorrect
entry, or a third-party re-upload. The problem therefore comes down to
how far a user without access to a model's internals can reconstruct,
without relying on self-reports, a tree that represents the
parent-child relationships.

Existing methods, however, have two limitations with respect to this
requirement (Section~\ref{sec:related}). First, methods that observe
only a model's output~\citep{nikolic2025model,phylolm} do not output a
directed tree. The output similarity these methods use is a
symmetric quantity that does not change when the two models are
swapped, so it cannot determine the direction of an edge. Second,
MoTHer~\citep{horwitz2025unsupervised}, which does output a directed
tree, determines edge direction from the weight distribution and so
presupposes white-box access, but the models to be verified include
ones whose weights are not public and that are available only through
a commercial API.

We therefore propose a method that reconstructs a directed tree without
referring to a model's weights, using as its only signal the
log-likelihood a model returns for a given input-output pair.
Fine-tuning raises the log-likelihood of the data used for training, on
that same data treated as an input-output pair. Because a child is
fine-tuned starting from its parent's parameters, we expect that, for
the data used in fine-tuning, the parent's increase in log-likelihood
is inherited by the child. Using this increase in log-likelihood, we
expect to be able to reconstruct a tree structure that represents a
model's derivation history, including the direction of derivation
introduced by fine-tuning.

This paper makes the following three contributions.

\begin{itemize}[leftmargin=*, topsep=0pt, itemsep=2pt, parsep=0pt]
  \item \textbf{Problem formulation and benchmark.} We formulate the
        problem of reconstructing a directed tree, including internal
        nodes, from only the teacher-forced log-likelihood and without
        referring to model weights, propose the two methods TRACE-Set
        and TRACE-Edge, and build a benchmark of 25 model trees across
        five configurations.
  \item \textbf{Experimental verification that fitting to training
          examples is inherited.} We show experimentally that
        fine-tuning causes a parent model's fit to its training
        examples to be inherited by its child, and that this can be
        observed when the training examples are used as probes. Using
        this property, we define a score function for estimating
        derivation relationships from fine-tuning.
  \item \textbf{Performance evaluation of directed-tree recovery with
          the proposed methods.} We show that TRACE-Edge, which uses
        none of the tree structure, the true parent, or any
        hyperparameter, achieves the best recovery of model derivation
        history under the held-out probe among all the methods we compare.
\end{itemize}

\section{Related Work}\label{sec:related}

\noindent
\textbf{Pairwise Derivation Testing.}
Model Provenance Testing (MPT)~\citep{nikolic2025model} tests whether a
model $g$ is derived from a candidate model $f$. It uses the similarity
between the next-token predictions of $f$ and $g$ as its test
statistic, and makes the decision through a test with a null
distribution formed from a set of control models known not to be in a
derivation relationship. This method operates under a weaker access
assumption than ours, black-box access, but its output only determines
whether an undirected derivation relationship exists for a pair of models, so it
does not recover a tree structure. The tree-recovery baseline we build
in this paper as an extension of MPT likewise outputs only an undirected
tree structure, and cannot determine edge direction without knowledge
of the root.

\noindent
\textbf{Output-Based Lineage Estimation.}
PhyloLM~\citep{phylolm} estimates a phylogenetic tree for a set of LLMs
by applying neighbor joining to a distance matrix defined from the
similarity of their generated outputs. This method also operates
under a weaker access assumption than ours, black-box access, but what
it outputs is an undirected dendrogram with no root specified, and
moreover only the leaf nodes are recovered, so it cannot recover a tree
structure that represents the derivation relationships.

\noindent
\textbf{Weight-Based Directed-Tree Recovery.}
MoTHer~\citep{horwitz2025unsupervised} is prior work that, like this
paper, formulates the task as recovering a directed model tree. It uses
weight distance to place edges and the kurtosis of the weight
distribution to orient them, and constructs a directed minimum spanning
tree with the Chu--Liu/Edmonds algorithm~\citep{chu1965shortest,edmonds1967optimum}.
It is the closest to this paper in that it outputs a directed tree,
including edge orientation, but white-box access is indispensable
because both edge placement and orientation refer to the weights.

Three additional lines of work share methodology or motivation with
ours: dataset inference, model fingerprinting, and watermarking. We
defer the differences between these works and our work to
Appendix~\ref{sec:related-appendix}.

\section{Preliminaries}\label{sec:prep}

\subsection{Language Models and Fine-Tuning}\label{sec:llm}

Let $V$ denote the vocabulary and $V^*$ the set of finite-length token
sequences over $V$. The LLMs considered in this paper are
autoregressive conditional language models with parameters $\theta$,
factorizing $p_\theta(y \mid x) = \prod_{i=1}^{|y|} p_\theta(y_i \mid
  x, y_{<i})$ for $x, y \in V^*$, where $y_{<i} = (y_1,\ldots,y_{i-1})$.
Supplying $y$ externally rather than from the model's own predictions
is teacher forcing; under it, $\log p_\theta(y_i \mid x, y_{<i})$ is a
deterministic quantity, independent of decoding strategy or
randomness, which a verifier can reproduce exactly. We write $m$ for a
model, $\theta_m$ for its parameters, and $p_m := p_{\theta_m}$.

Let a dataset for task $\tau$ be
$D_\tau = \{(x_j, y_j)\}_j \subset V^* \times V^*$. Fine-tuning takes
existing parameters $\theta$ as the initial value and minimizes
$-\sum_{(x,y) \in D_\tau} \log p_\theta(y \mid x)$ to obtain
$\theta' = \mathrm{FT}(\theta, D_\tau, c)$, where $c$ is the training
configuration (optimizer, learning rate, number of epochs). Throughout
this paper, $c$ is held fixed across all
nodes of a tree (Section~\ref{sec:tree-setup}). This one-directional
operation from $\theta$ to $\theta'$ defines a
parent-child relationship, and is the only derivation relationship this
paper treats. It covers both full-parameter fine-tuning and
parameter-efficient methods such as LoRA~\citep{lora}, since
either yields a single $\theta'$ from a single $\theta$. By contrast,
model merging, which combines the parameters of two or
more parents, does not fit a tree-structured framework and is outside
the scope of this work.

\subsection{Model Trees}\label{sec:tree}

When a model is repeatedly fine-tuned, taking each result as the
parent of the next, the derivation relationships form a tree
structure, which we call a model tree $T = (M, E)$. Here $M$ is the
set of models, and $E \subset M \times M$ is the set of directed edges
representing fine-tuning from parent to child. The root $r \in M$ is a
pretrained model that has not been fine-tuned, and every non-root node
$m \in M \setminus \{r\}$ has exactly one parent $\mathrm{pa}(m)$. Each
node $m \neq r$ is the result of fine-tuning on a single task $\tau_m$
starting from its parent's parameters, that is,
$\theta_m = \mathrm{FT}(\theta_{\mathrm{pa}(m)}, D_{\tau_m}, c)$. We
call the nodes other than $m$ on the path from the root to $m$ the
ancestors of $m$, and a node that shares $m$'s parent but is not on the path from the root to $m$
a sibling of $m$. A model tree is a directed tree in which every node is reachable
from $r$. This paper assumes that $M$ contains every node of $T$ without
omission, that is, that no unobserved intermediate-generation models
exist.

\subsection{Verifier Setting}\label{sec:threat}

In this section, we call the party who performs lineage verification the
\emph{verifier}, and define, in turn, the verifier's situation, the
verifier's available access, and the verifier's prior knowledge assumed.

\noindent
\textbf{Setting.}
We primarily consider a verifier who is a user seeking to adopt a
public model in their own system. The verifier observes a set of
models $M$ published on a model-sharing platform such as Hugging Face.
Each model in $M$ is provided in a form that supports inference, but
the derivation relationships $E$ among them are either unreported, or
reported without any guarantee of correctness.

\noindent
\textbf{Verifier's Available Access.}
For any $m \in M$, the verifier can supply any $(x,y) \in V^* \times
  V^*$ and obtain the per-token log-likelihoods
$\log p_m(y_i \mid x, y_{<i})_{i=1}^{|y|}$. This setting is intended to
model inference backends that return log-likelihoods (such as vLLM~\citep{kwon2023efficient} and
Text Generation Inference~\citep{huggingface_tgi}) or inference APIs that return logprobs. It
is stronger than a fully black-box setting that returns only generated
output, and weaker than a white-box setting that requires the
parameters $\theta_m$, so we call it gray-box access in this paper. The
verifier has no access to the weights $\theta_m$, gradients, training
logs, or intermediate-layer representations, and cannot obtain any
cooperation from the provider.

\noindent
\textbf{Verifier's Prior Knowledge.}
The verifier knows the root node $r$ of $M$; Appendix~\ref{sec:appendix-rootest} relaxes this
assumption by providing a procedure for estimating an unknown root.
We assume a \emph{closed-world setting}: the verifier knows a finite
task set $\mathcal{T}=\{\tau_1,\ldots,\tau_T\}$ covering all fine-tuning
tasks along the model tree, i.e., $\{\tau_m \mid m \in M \setminus \{r\}\} \subseteq \mathcal{T}$.
This assumption is reasonable for auditing and intellectual property verification when training
records or sufficiently detailed model cards disclose the relevant tasks,
while the derivation relationships remain unknown.
We leave the open-world setting with incomplete task knowledge to future work,
focusing here on provenance inference without the
additional uncertainty introduced by unknown tasks.

For each task $\tau \in \mathcal{T}$, we suppose the verifier can obtain a
\emph{probe set} $P_\tau$. We consider two cases: \emph{train},
where the examples themselves are available
(e.g., auditing outsourced fine-tuning or checking license compliance),
and \emph{held-out}, where the examples are drawn from the same distribution
but were not used for training (e.g., tracing the inheritance of vulnerabilities or safety risks).

\subsection{Goal and Evaluation Metrics}\label{sec:problem}

\noindent
\textbf{Goal.}
Given a node set $M$, a root $r$, and a task set $\mathcal{T}$, with
gray-box access to the models in the node set, the verifier's goal is
to recover the model tree $T=(M,E)$. For each non-root node
$m \in M \setminus \{r\}$, the verifier outputs a parent estimate
$\widehat{\mathrm{pa}}(m) \in M \setminus \{m\}$. The verifier's goal
is for the resulting edge set
$\hat{E} = \{ (\widehat{\mathrm{pa}}(m), m) \}_{m \neq r}$ to match the
true edge set $E$. Unlike a pairwise formulation that separately
decides whether two models are in a derivation relationship, this
formulation simultaneously requires (i) determining the direction of
each edge and (ii) recovering the whole tree, including internal
nodes, consistently.

\noindent
\textbf{Evaluation Metrics.}
We measure recovery performance with two metrics. First, we define the
parent-match accuracy as
\begin{equation}
  \mathrm{Acc}(\hat{E}) = \frac{1}{|M| - 1}
  \bigl| \{\, m \in M \setminus \{r\} : \widehat{\mathrm{pa}}(m) = \mathrm{pa}(m) \,\} \bigr|
  \label{eq:acc}
\end{equation}
Each node has $|M|-1$ candidate parents, namely every other node, so
the expected value under a random choice of parent, that is, the
chance level, is $1/(|M|-1)$. This is $\approx 0.167$ for the 7-node
trees used in this paper and $\approx 0.125$ for the 9-node tree.

Next, we define the depth-match accuracy. For each $m \in M$, the path
from the root is unique, and we call the number of edges on it the
depth $d_T(m)$ of $m$, so that $d_T(r)=0$ and
$d_T(m)=d_T(\mathrm{pa}(m))+1$ for $m\neq r$.

We also restrict the verifier's output $\hat{T}=(M,\hat{E})$ to a
spanning directed tree rooted at $r$ (every method considered in this
paper satisfies this). We then define the depth-match accuracy as
\begin{equation}
  \mathrm{DAcc}(\hat{T}) = \frac{1}{|M|-1}
  \bigl|\{\, m \in M\setminus\{r\} : d_{\hat{T}}(m) = d_T(m) \,\}\bigr|
  \label{eq:dacc}
\end{equation}
The root is excluded from the denominator because it always matches at
depth 0. The chance level is $\approx 0.333$ for the 7-node trees used
in this paper and $\approx 0.250$ for the 9-node tree.

\section{Method}\label{sec:method}

This section proposes a method for recovering a directed tree from
gray-box access alone. At its core is the trace that fine-tuning
history leaves in a model's output. The fit to a fine-tuning task
applied somewhere along the path from the root $r$ to a non-root node
$m$ can be observed as a change in the log-likelihood $m$ assigns to
the training data used for that fine-tuning. Our method formalizes
this parent-child difference in log-likelihood as a per-task
deviation. This deviation reflects a one-directional property of the
operation. Fine-tuning from parent to child tends to raise a task's
log-likelihood, so the deviation can be used to determine the
direction of a parent-child relationship.

Based on this observation, this section proposes two methods under the
name \textbf{TRACE} (Teacher-forced Reconstruction of Ancestry from
Conditional-likelihood Evidence): \textbf{TRACE-Set}
(Section~\ref{sec:m8}), which determines parents from the inclusion
relationships among task sets, and \textbf{TRACE-Edge}
(Section~\ref{sec:m9}, the main method of this paper), which exactly
solves for the maximum spanning directed tree using the per-task
deviation as edge weights.

\subsection{Deviations}\label{sec:deltas}

This section formalizes the deviation that TRACE-Set and TRACE-Edge
use in common. We define the sequence-averaged log-likelihood with which model $m$
generates the target output $y$ given input $x$ as
\begin{equation}
  S_m(x,y) = \frac{1}{|y|} \sum_{i=1}^{|y|} \log p_m(y_i \mid x, y_{<i})
  \label{eq:score}
\end{equation}
For $(x,y) \in P_\tau$ in the probe example set of task
$\tau \in \mathcal{T}$, $S_m(x,y)$ in Eq.~\ref{eq:score} measures how
well $m$ fits $\tau$. If $m$ was fine-tuned on $\tau$ while deriving
from its parent $p$, we expect $S_m(x,y)$ to exceed $S_p(x,y)$ by more
than it would if $m$ had not been fine-tuned on $\tau$. Given the
example set $P_\tau \subset \mathcal{V}^* \times \mathcal{V}^*$ of task
$\tau$, we therefore define the deviation $\delta_\tau(f \to m)$ of
model $m$ relative to a reference model $f$ as
\begin{equation}
  \delta_\tau(f \to m) = \frac{1}{|P_\tau|} \sum_{(x,y) \in P_\tau}
  \bigl( S_m(x,y) - S_f(x,y) \bigr) \in \mathbb{R}
  \label{eq:delta-task}
\end{equation}
That is, $\delta_\tau(f \to m)$ shows the change in fit to task $\tau$
of $m$ relative to $f$. When $\delta_\tau(f \to m)$ is large, we
consider it likely that $\tau$ was used to fine-tune $m$ somewhere
along its derivation from $f$.

Recall that $D_\tau$ (Section~\ref{sec:llm}) is the fine-tuning set.
For the two probe conditions of Section~\ref{sec:threat}, we write
$P_\tau^{\text{train}} \subset D_\tau$ and $P_\tau^{\text{held-out}}$,
the latter drawn from the same distribution as $D_\tau$ but disjoint
from it.

\subsection{TRACE-Set}\label{sec:m8}

This section defines TRACE-Set, a method that recovers the tree
structure from inclusion relationships among tasks. It estimates, for
each node, the set of tasks fine-tuned along the path from the root,
and identifies the parent-child relationships, that is, the tree
structure, from the inclusion relationships among these sets.

If a task $\tau \in \mathcal{T}$ was used for fine-tuning somewhere
along the path from the root $r$ to $m$, we expect $\delta_\tau(r \to
  m)$ to take a positive value. For a task not on the path, conversely,
this increase does not occur. We therefore expect that looking at
which tasks give a positive deviation relative to $r$ lets us estimate
the history of tasks that $m$'s ancestors used for fine-tuning.

For each node $m$, we therefore define the ancestor task set, the set
of tasks estimated to fit substantially better than the root, as
\begin{equation}
  A(m) = \{\, \tau \in \mathcal{T} : \delta_\tau(r \to m) > \eta \,\}
  \label{eq:ancestor-set}
\end{equation}
Here $\eta$ is a positive threshold, the only hyperparameter TRACE-Set
has. We describe how we choose its value in Section~\ref{sec:tree-setup}.

Because the true parent $p$ of $m$ is the ancestor whose path ends one
task before $m$'s, we expect $A(m)$ to contain $A(p)$ with a
difference of exactly one task. That is, we consider $p$ to satisfy
\begin{equation}
  A(f) \subsetneq A(m), |A(m) \setminus A(f)| = 1  \label{eq:parent-condition}
\end{equation}

TRACE-Set estimates the parent of $m$ as an $f$ that satisfies
Eq.~\ref{eq:parent-condition}. Because of the thresholding by $\eta$,
no such $f$ may exist, or several may. In the former case, we relax
the condition to the candidate with the smallest set difference. In
the latter, we choose the $f$ with the largest deviation among the
tasks in the set difference. We summarize the full procedure in
Algorithm~\ref{alg:dset} (Appendix~\ref{sec:appendix-algo}).\footnote{Without the inclusion-based
  filtering, directly choosing the $f$ with the largest deviation among
  the tasks in the set difference misidentifies the true parent whenever
  an ancestor's fine-tuning task has a larger deviation than the node's
  own fine-tuning task, because an ancestor with a larger set difference
  then gets chosen instead. In practice, we observed cases where the
  deviation for gsm8k, which is far larger than that of other tasks,
  exceeded the deviation for the node's own fine-tuning task.}

\subsection{TRACE-Edge}\label{sec:m9}

This section defines TRACE-Edge, a method that introduces an
asymmetric edge weight for each ordered pair of models and finds the
directed tree that maximizes the sum of edge weights over the whole
tree. Whereas TRACE-Set determines parent-child relationships locally
at each node, TRACE-Edge determines the parents of all nodes
simultaneously, using a complete directed graph in which models are
nodes and parent-child relationships are directed edges. We define, as
the weight of each directed edge, a score function that grows larger
the more likely a derivation relationship is, and estimate the tree
structure from these scores.

\noindent
\textbf{Designing the Score Function.}
The core idea of this method is that, when the child's deviation is
measured relative to its true parent, the deviation appears only for
the single task used to fine-tune the child itself. When a candidate
parent model $f$ is the true parent of $m$, $m$ inherits the whole of
$f$'s fine-tuning history and is then further fine-tuned only on task
$\tau_m$. We should therefore expect $\delta_{\tau_m}(f \to m)$ to
take a large positive value, and the deviations $\delta_{\tau'}(f \to
  m)$ for the other tasks $\tau' \neq \tau_m$ to be close to 0.

In what follows, given $f$ and $m$, we design a score function $s$
that judges whether $f$ is the parent of $m$. The key point is that
$\delta_\tau(f \to m)$ changes substantially only for tasks fine-tuned
somewhere on the path between $f$ and $m$, and is positive when that
path is traversed from parent to child and negative when traversed
from child to parent. We consider four cases according to the
relationship between $f$ and $m$: (1) $f$ is the true parent of $m$,
(2) $f$ is an ancestor of $m$ other than its parent, (3) $f$ is a
descendant of $m$, and (4) $f$ is neither an ancestor nor a descendant
of $m$. These four cases cover every $f$ other than $m$ itself.

\noindent
\textbf{(1) $f$ is the parent of $m$.}
$\delta_\tau(f \to m)$ is positive only for the task $\tau_m$ used to
fine-tune $f$ into $m$.

\noindent
\textbf{(2) $f$ is an ancestor of $m$ other than its parent.}
$m$ is fine-tuned on two or more tasks along the path from $f$. So
there are two or more $\tau$ for which $\delta_\tau(f \to m)$ is
positive.

\noindent
\textbf{(3) $f$ is a descendant of $m$.}
$f$ is fine-tuned on one or more tasks along the path from $m$. So
there is at least one $\tau$ for which $\delta_\tau(f \to m)$ is
negative.

\noindent
\textbf{(4) $f$ is neither an ancestor nor a descendant of $m$.}
The paths from the root to $f$ and to $m$ diverge, so let $a$ be their
point of divergence, the deepest common ancestor of $f$ and $m$. The
path from $f$ to $m$ then splits into two parts: from $f$ to $a$,
going from child to parent, and from $a$ to $m$, going from parent to
child. The former can be treated as in (3) and the latter as in (1)
and (2), and since there are tasks fine-tuned along each part of the
path, there is at least one $\tau$ for which $\delta_\tau(f \to m)$ is
positive and at least one for which it is negative.

Taking cases (1) through (4) together, only when $f$ is the true
parent of $m$ does $\delta_\tau(f \to m)$ take a large positive value
for one task and a value close to 0 for the others. Fix an ordered
pair $(f,m)$ for $f, m \in M$, and write the deviations
$\delta_\tau(f \to m)$ for $\tau \in \mathcal{T}$ sorted in descending
order as $\delta_{(1)} \geq \delta_{(2)} \geq \ldots \geq \delta_{\min}
  = \min_{\tau\in\mathcal{T}} \delta_{\tau}(f \to m)$. We then define the
score indicating that $f$ is the true parent of $m$ as
\begin{equation}
  s(f \to m) = \delta_{(1)} - \max\bigl(\delta_{(2)},\ |\delta_{\min}|\bigr)
  \label{eq:m9-weight}
\end{equation}
The first term of $s$ takes the largest positive deviation, and the
second term takes either the second-largest positive deviation or the
smallest (most negative) deviation.

Table~\ref{tab:score-cases} summarizes the behavior of each term and
of $s(f \to m)$ for cases (1) through (4). Here $\mathcal{T}^+ =
  \{\tau \in \mathcal{T} \mid \delta_\tau(f \to m) > \epsilon\}$ and
$\mathcal{T}^- = \{\tau \in \mathcal{T} \mid \delta_\tau(f \to m) < -
  \epsilon\}$, where $\epsilon$ is a small positive value representing
the error in the deviation. Only in case (1) is $s(f \to m)$ composed
of positive terms alone, so we expect it to take a larger positive
value than in the other cases.

\begin{table}[t]
  \centering
  \caption{Behavior of $\delta_\tau(f \to m)$ and the score $s(f \to m)$ for each relationship between $f$ and $m$}
  \label{tab:score-cases}
  \begin{tabular}{lcccccc}
    \toprule
    Case & $|\mathcal{T}^+|$ & $|\mathcal{T}^-|$ & $\delta_{(1)}$ & $\delta_{(2)}$ & $\delta_{\min}$ & $s(f \to m)$                     \\
    \midrule
    (1)  & $1$               & $0$               & $>\epsilon$    & $\approx 0$    & $\approx 0$     & $>\epsilon$                      \\
    (2)  & $\geq2$           & $0$               & $>\epsilon$    & $>\epsilon$    & $\approx 0$     & $\delta_{(1)} - \delta_{(2)}$    \\
    (3)  & $0$               & $\geq1$           & $\approx 0$    & $\approx 0$    & $< -\epsilon$   & $< -\epsilon$                    \\
    (4)  & $\geq1$           & $\geq1$           & $>\epsilon$    & ---            & $< -\epsilon$   & $\delta_{(1)} - |\delta_{\min}|$ \\
    \bottomrule
  \end{tabular}
\end{table}
\vspace{-4pt}

Computing the score of Eq.~\ref{eq:m9-weight} as the edge weight for
every ordered pair $(f,m)$ gives a complete directed graph on the
vertex set $M$. On the graph obtained by removing every edge into $r$,
TRACE-Edge exactly finds, with the
Chu--Liu/Edmonds algorithm~\citep{chu1965shortest,edmonds1967optimum},
the spanning directed tree rooted at $r$ that maximizes the sum of edge
weights. It imposes no constraint on the number of children or the
depth, and has no hyperparameter. We summarize the procedure in
Algorithm~\ref{alg:trace-edge} (Appendix~\ref{sec:appendix-algo}).

\section{Experiments}\label{sec:eval}

\subsection{Experimental Setup}\label{sec:tree-setup}

\begin{wrapfigure}{r}{0.4\textwidth}
  \centering
  \begin{tikzpicture}[
      mn/.style={draw, rounded corners=1.5pt, inner sep=1.2pt,
          font=\scriptsize, minimum height=3.4mm, minimum width=7mm},
      rt/.style={mn, fill=black!10},
      ed/.style={-latex, semithick},
      el/.style={font=\tiny, midway, above, sloped, inner sep=1pt}
    ]
    \node[rt] (r)  at (0,-1.5)   {$r$};
    \node[mn] (gm) at (1.75,-0.3) {$g_{\mathrm{math}}$};
    \node[mn] (gs) at (1.75,-2.1) {$g_{\mathrm{sum}}$};
    \node[mn] (c1) at (3.7,0)     {$c_1$};
    \node[mn] (c2) at (3.7,-0.6)  {$c_2$};
    \node[mn] (c3) at (3.7,-1.5)  {$c_3$};
    \node[mn] (c4) at (3.7,-2.7)  {$c_4$};
    \draw[ed] (r)  -- (gm) node[el] {math};
    \draw[ed] (r)  -- (gs) node[el] {summ};
    \draw[ed] (gm) -- (c1) node[el] {data2text};
    \draw[ed] (gm) -- (c2) node[el] {translation};
    \draw[ed] (gs) -- (c3) node[el] {data2text};
    \draw[ed] (gs) -- (c4) node[el] {translation};
  \end{tikzpicture}
  \caption{Structure and task assignment of HET-3B. Edge labels denote
    the task each node used for fine-tuning. Task details and the
    structure of all five trees are shown in Figure~\ref{fig:trees-all}
    in Appendix~\ref{sec:tree-detail}.}\label{fig:trees}
  \vspace{-10pt}
\end{wrapfigure}
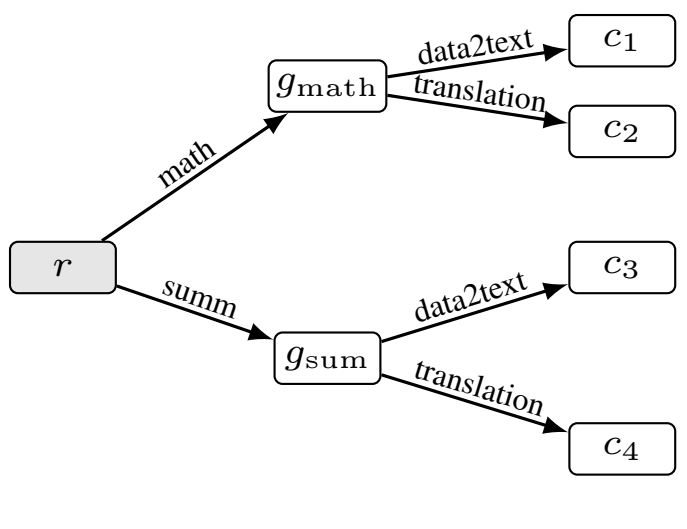

\noindent
\textbf{Model Trees.}
For evaluation, we use five model trees that we constructed ourselves.
The reference tree is a 7-node, depth-2 tree rooted at
Llama-3.2-3B~\citep{llama3}, which we call HET-3B. Figure~\ref{fig:trees}
shows the structure of HET-3B. Starting from the root Llama-3.2-3B,
shown as $r$ in Figure~\ref{fig:trees}, HET-3B has two nodes
$g_{\mathrm{math}}$ and $g_{\mathrm{sum}}$ fine-tuned on a math
word-problem task and a summarization task, from which
four nodes $c_1, c_2, c_3, c_4$ derive, each child fine-tuned on one of
two tasks: data-to-text generation and translation.

\noindent
\textbf{Tree Variants.}
The remaining four trees each change exactly one element of HET-3B:
HET-8B replaces the root with Llama-3.1-8B to evaluate the effect of
model size; HET-QWEN replaces the root with Qwen2.5-3B to evaluate the
effect of model family; DEEP is a depth-3 tree where two further nodes
$d_1, d_2$, fine-tuned on two summarization tasks, derive from node $c_3$ of HET-3B, to evaluate the
effect of tree depth; and HOMO keeps HET-3B's structure but changes
every node's fine-tuning task to a summarization task, to examine
recovery performance when the fine-tuning tasks resemble each other.
A detailed list of all five trees is given in Table~\ref{tab:trees} in
Appendix~\ref{sec:appendix-trees}, and the models and datasets used
are given in Appendix~\ref{sec:appendix-data}.

For Section~\ref{sec:recover-eval}, we fine-tune every node with
LoRA~\citep{lora}, using hyperparameters shared across all tasks and
trees (values in Appendix~\ref{sec:appendix-hp}). For each of the five
configurations, we construct five trees with different random seeds,
for 25 trees in total, and report every metric below as the mean and
standard deviation over the 5 seeds for each configuration. To check that
reliance on LoRA does not bias the results, we ran the same evaluation
on one additional tree fine-tuned with all parameters updated (Full
FT). Appendix~\ref{sec:appendix-fullft} reports that it shows the same trend under train as the five LoRA trees.

\noindent
\textbf{TRACE-Set's Threshold.}
We choose the threshold $\eta$ in Eq.~\ref{eq:ancestor-set}
independently for each configuration by leave-one-tree-out (candidate
values and the resulting per-configuration choices are given in
Appendix~\ref{sec:appendix-hp}).

\noindent
\textbf{Probe Sets.}
To compute the deviation $\delta_\tau$, we use $P_\tau^{\text{train}}$
and $P_\tau^{\text{held-out}}$ as defined in Section~\ref{sec:deltas},
with $100$ examples per task across the seven fine-tuning tasks used in
the five trees ($700$ examples total, common to every tree and node;
task details in Appendix~\ref{sec:appendix-data}).

\noindent
\textbf{Baselines.}
As a baseline, we use the trivial prediction that predicts every node
as a direct child of the root $r$ (the star baseline). We also add, as
a comparison method, an MST (Minimum Spanning Tree) estimated from
MPT's~\citep{nikolic2025model} similarity, using the output mismatch
rate (the Top-1 prediction disagreement rate) as a symmetric distance.
MST is a black-box method that measures difference from the model's
output alone, and so operates under a weaker assumption than the
gray-box access required by TRACE-Set and TRACE-Edge. Because it finds
an undirected tree that minimizes a symmetric distance, however, MST
does not output edge direction. We therefore compute each
evaluation metric on a tree oriented from root to leaves under the
additional assumption that the root is known (Section~\ref{sec:recover-eval}).
We compare these methods with our proposed methods.

We also consider a white-box MST that uses the $L_2$ distance of the
parameter update $\Delta\theta$ as its edge weight (corresponding to
an existing method that relies on weight-space
distance~\citep{horwitz2025unsupervised}). Because it requires the
weights of every model, its access conditions differ substantially
from this paper's gray-box assumption. We do not include it in the
main results table, and mention it only as a reference value in the
text.

\noindent
\textbf{Evaluation Metrics.}
Our main metrics are the parent-match accuracy Acc (Eq.~\ref{eq:acc})
and the depth-match accuracy DAcc (Eq.~\ref{eq:dacc}). We compare the
estimated edge set directly with the true edge set.

\subsection{Validity of the Deviation}\label{sec:delta-eval}

This section confirms, for both $P_\tau^{\text{train}}$ and
$P_\tau^{\text{held-out}}$, that the deviation is positive for tasks
on the path from the root to $m$ and not positive for tasks not on that
path. Taking as the unit of analysis each cell of the deviation matrix that
arranges the deviation $\delta_\tau(r \to m)$ for every node and task (32
non-root nodes $\times$ 7 tasks, details in
Appendix~\ref{sec:appendix-probe}), the sign of the deviation
correctly identifies the task used for fine-tuning in 193 of 224 cells
($0.862$) when the probe example set uses training examples.
Restricted to the ancestor-task cells, which directly determine parent
identification, this rate is even higher, at $22$ of $24$ cells
($0.917$). With the held-out probe, the deviation for each node's own fine-tuning
task still stays positive in 25 of 32 cells (0.781), well above the
0.500 expected from a random sign, while the ancestor cells fall to
0.458 as fit to the training examples themselves is lost. We
consider this drop in the estimation rate to be one cause of the
recovery-performance decline from train to held-out discussed below
(Section~\ref{sec:recover-eval}). We revisit this point in
Section~\ref{sec:limitation}.

\subsection{Main Results}\label{sec:recover-eval}

This section compares TRACE-Set and TRACE-Edge against the black-box
MST and the star baseline, under the train and held-out probe
conditions, with the true root supplied.\footnote{The white-box MST
  using parameter distance as its edge weight achieves an
  undirected-edge recall of 1.000 on every tree. This is a reference
  value for the case where the weights of every model are available,
  and it is not usable under the gray-box condition set out in
  Section~\ref{sec:threat}.}

\begin{table*}[t]
  \caption{Per-tree parent-match accuracy, and overall parent-match
    accuracy (Acc) and depth-match accuracy (DAcc) across all five
    trees, with the true root supplied as input. The per-tree columns
    are the mean over 5 seeds. The Acc and DAcc columns are the mean
    $\pm$ standard deviation over all 25 runs (5 trees $\times$ 5
    seeds). The access column is Black for black-box output, Gray for
    gray-box output, and --- for no access to the model. Bold marks
    the best value within the same probe. Results with an estimated
    root are given in Appendix~\ref{sec:appendix-rootest-recovery}.
  }
  \label{tab:main}
  \begin{center}
    \small
    \resizebox{\linewidth}{!}{%
      \begin{tabular}{lllccccccc}
        \toprule
              &        &               & \multicolumn{5}{c}{Per-tree parent-match accuracy} & \multicolumn{2}{c}{All 5 trees}                                                                                                  \\
        \cmidrule(lr){4-8} \cmidrule(lr){9-10}
        probe & Access & Method        & HET-3B                                             & HET-8B                          & HET-QWEN       & DEEP           & HOMO           & Acc                  & DAcc                 \\
        \midrule
        \multirow{4}{*}{train}
              & ---    & Star baseline & 0.333                                              & 0.333                           & 0.333          & 0.250          & 0.333          & $0.317\pm0.034$      & $0.317\pm0.034$      \\
              & Black  & MST           & 0.700                                              & 0.667                           & 0.633          & 0.600          & 0.500          & $0.620\pm0.165$      & $0.553\pm0.189$      \\
              & Gray   & TRACE-Set     & \textbf{0.833}                                     & \textbf{1.000}                  & \textbf{0.900} & 0.650          & \textbf{0.800} & $\bm{0.837\pm0.165}$ & $\bm{0.857\pm0.153}$ \\
              & Gray   & TRACE-Edge    & 0.733                                              & 0.667                           & 0.833          & \textbf{0.800} & 0.667          & $0.740\pm0.084$      & $0.707\pm0.129$      \\
        \midrule
        \multirow{4}{*}{held-out}
              & ---    & Star baseline & 0.333                                              & 0.333                           & 0.333          & 0.250          & 0.333          & $0.317\pm0.034$      & $0.317\pm0.034$      \\
              & Black  & MST           & 0.633                                              & \textbf{0.567}                  & 0.400          & 0.500          & 0.433          & $0.507\pm0.128$      & $0.468\pm0.081$      \\
              & Gray   & TRACE-Set     & 0.533                                              & 0.333                           & 0.400          & 0.500          & \textbf{0.733} & $0.500\pm0.152$      & $\bm{0.665\pm0.133}$ \\
              & Gray   & TRACE-Edge    & \textbf{0.667}                                     & 0.233                           & \textbf{0.600} & \textbf{0.625} & 0.667          & $\bm{0.558\pm0.206}$ & $0.618\pm0.174$      \\
        \bottomrule
      \end{tabular}}
  \end{center}
\end{table*}
\vspace{-4pt}

Table~\ref{tab:main} shows the main results. Under train, TRACE-Edge
and TRACE-Set both outperform MST, with a parent-match accuracy of
$0.837$ for TRACE-Set, $0.740$ for TRACE-Edge, and $0.620$ for MST.
Under held-out, the advantage of our two proposed methods shrinks: the
parent-match accuracy is $0.558$ for TRACE-Edge, $0.507$ for MST, and
$0.500$ for TRACE-Set, with TRACE-Edge still best but by a
smaller margin than under train.

\section{Limitations}\label{sec:limitation}

\noindent
\textbf{The Known-Root Assumption.}
We constructed TRACE-Set and TRACE-Edge under the assumption that the
root $r$ is known, and proposed a root-estimation method to relax it
(Appendix~\ref{sec:appendix-rootest}). Though this approach
achieves a reasonable root-estimation accuracy,
a more effective root-estimation method, or building a method that does not
assume the root is known, is future work.

\noindent
\textbf{The Train-Probe-availability Assumption.}
The performance of our method depends on the availability of the
training examples themselves. As the main results
(Section~\ref{sec:recover-eval}) show, TRACE-Edge's parent-match
accuracy falls from $0.740$ under train, which probes with the
training examples, to $0.558$ under held-out, which probes with
non-training examples from the same distribution.

The cause lies in the nature of the fit that the teacher-forced
log-likelihood measures. Because fine-tuning is an operation that
raises the log-likelihood of the prompts used for training, using the
training examples themselves for teacher forcing yields a strong
signal in the deviation that is closer to fit to the individual data
than to generalized fit to the task or domain. Under held-out, this
signal weakens, and the rate of correctly estimating ancestor tasks
drops by roughly half, from $0.917$ to $0.458$
(Section~\ref{sec:delta-eval}). Fit to a node's own fine-tuning task is
relatively preserved even under held-out, whereas the estimation rate
for the ancestor cells needed to identify the parent drops first, and
this translates directly into the decline in parent-match accuracy.

For the use cases discussed in Section~\ref{sec:threat} where the
verifier cannot obtain the training examples themselves, held-out is
the realistic probe condition, so this decline needs to be addressed.
Resolving its cause calls for constructing probes that respond only to
the fine-tuning of a specific task without using the training examples
themselves.

\section{Conclusion}

This paper addressed the problem of recovering an LLM's derivation
relationships as a directed tree, using only the deviation of the
teacher-forced log-likelihood and without referring to the model's
weights. We showed that the deviation is an asymmetric quantity
defined for each pair of models and can determine the direction of an
edge, and proposed TRACE-Edge, a method that recovers the maximum
spanning directed tree using the per-task deviation as edge weights.
In an evaluation on a benchmark of 25 model trees across five
configurations, TRACE-Edge, which uses none of the tree structure, the
true parent, or any hyperparameter, achieved the best parent-match
accuracy among all the methods we compared under the held-out probe
condition, which uses only non-training examples from the same
distribution as the training examples. At the same time, recovery
performance declines when the training examples used for fine-tuning
are not available, meaning that what the deviation captures is not
task competence but fit to the training examples themselves;
constructing probes that respond only to the fine-tuning of a specific
task without using the training examples is the main direction for
future work.

\subsection*{AI use statement}

We used generative AI (Claude) for implementing code and for
assisting with translation, both tasks with
required disclosure. We did not use generative AI to develop the
method, formulate the theoretical claims, design the experiments, or
interpret the results, and this work involves no synthetic dataset
generation, no proofs, and no qualitative or thematic data analysis,
so those required-disclosure categories do not apply. We also used
generative AI for editing the translated text for readability, for
creating and modifying figures, for literature search, for suggesting
the paper's structure, and for checking consistency between the main
text and its tables and references, all tasks with recommended
disclosure.

We reviewed all AI-assisted work. AI-generated code was tested and
checked by the authors, including comparison against ground truth and
sanity checks against random baselines. AI-modified figures were
checked against the underlying source data. Papers found through
AI-assisted literature search were verified to be real, existing
publications by checking them against arXiv, ACL Anthology, NeurIPS
proceedings, and OpenReview.
AI-suggested structural changes and
consistency checks between the main text and its tables and references
were reviewed by the authors against the original manuscript before
being applied. We take responsibility for the final content of this
work, including text, claims, and artifacts produced with the aid of
generative AI.

\subsection*{Ethics statement}

This work uses publicly available pretrained language models and
publicly available datasets, and involves no human subjects and no
personally identifiable data. The verification method we propose is
intended to help model users and providers confirm reported lineage
relationships, and a party able to run our verifier already has the
gray-box access our method requires, so we are not aware of a harmful
dual use of this capability. We do not identify any other ethical
concerns specific to this work.

\subsection*{Reproducibility statement}

TRACE-Set and TRACE-Edge are formally defined in
Section~\ref{sec:method}, and Algorithms~\ref{alg:dset}
and~\ref{alg:trace-edge} give complete pseudocode for both. The
experimental setup, including the model-tree configurations and the
evaluation protocol, is described in
Section~\ref{sec:tree-setup}, with the full list of base models and
datasets in Appendix~\ref{sec:appendix-data} and the training
configuration in Appendix~\ref{sec:appendix-hp}. We may release the
benchmark of model trees constructed for this evaluation as an
anonymous public resource in the future.

\bibliography{ref}
\bibliographystyle{iclr2027_conference}

\appendix

\section{Additional Related Work}\label{sec:related-appendix}

This section surveys research areas closely related to this paper's
problem setting.

\noindent
\textbf{Dataset-Level Membership Determination.}
LLM Dataset Inference~\citep{maini2024llm} proposes dataset inference,
which determines whether an entire collection of documents, such as a
book, rather than individual examples, was used in training. It
selects and aggregates log-likelihood-based signals from several
membership inference methods and reaches a decision through a
statistical test, which is methodologically close to how this paper
uses the log-likelihood deviation (the deviation $\bm{\delta}$
formalized in Section~\ref{sec:method}) as its signal. The goal,
however, is fundamentally different. That method asks the binary
question of whether a specific dataset was used to train a given
model, and does not address the lineage relationship between models or
its direction.

\noindent
\textbf{Model Fingerprinting and Watermarking.}
Other means of tracking a model's origin include model fingerprinting,
which identifies the base model from properties of its
parameters~\citep{zeng2024huref}, and watermarking, in which the
provider embeds a signal in the
output~\citep{peng2023you}. These methods
aim to identify which base model or provider a model derives from, and
require by design that the signal survive subsequent training or
modification. Both also presuppose that the model owner computes the
fingerprint or embeds the watermark. This paper, by contrast, assumes
no cooperation from the owner, and recovers a directed tree, including
parent-child relationships, from the changes introduced at each stage
of fine-tuning.

\section{Algorithms}\label{sec:appendix-algo}

This section gives the full procedures for TRACE-Set
(Algorithm~\ref{alg:dset}, referenced from Section~\ref{sec:m8}) and
TRACE-Edge (Algorithm~\ref{alg:trace-edge}, referenced from
Section~\ref{sec:m9}).

\begin{algorithm}[H]
  \caption{TRACE-Set}
  \label{alg:dset}
  \begin{algorithmic}[1]
    \REQUIRE Node set $M$, root node $r \in M$, per-task deviations $\{\delta_\tau(r \to m)\}_{m \in M,\, \tau \in \mathcal{T}}$, threshold $\eta$
    \ENSURE Parent function $\mathrm{pa} : M \setminus \{r\} \to M$
    \STATE $A(r) \gets \emptyset$
    \FOR{$m \in M \setminus \{r\}$}
    \STATE $A(m) \gets \{\, \tau \in \mathcal{T} : \delta_\tau(r \to m) > \eta \,\}$
    \ENDFOR
    \FOR{$m \in M \setminus \{r\}$}
    \STATE $S \gets \{\, f \in M \setminus \{m\} : A(f) \subsetneq A(m) \,\}$
    \IF{$S = \emptyset$}
    \STATE $\mathrm{pa}(m) \gets r$ \COMMENT{Occurs only when $A(m) = \emptyset$}
    \STATE \textbf{continue}
    \ENDIF
    \STATE $k \gets \min_{f \in S} |A(m) \setminus A(f)|$
    \STATE $C \gets \{\, f \in S : |A(m) \setminus A(f)| = k \,\}$
    \STATE $\mathrm{pa}(m) \gets \argmax_{f \in C} \ \max_{\tau \in A(m) \setminus A(f)} \delta_\tau(r \to m)$ \COMMENT{Ties broken uniformly at random}
    \ENDFOR
    \RETURN $\mathrm{pa}$
  \end{algorithmic}
\end{algorithm}

\begin{algorithm}[H]
  \caption{TRACE-Edge}
  \label{alg:trace-edge}
  \begin{algorithmic}[1]
    \REQUIRE Node set $M$, root $r$, per-task deviations $\{\delta_\tau(f\to m)\}_{f,m \in M,\, f \neq m,\, \tau \in \mathcal{T}}$
    \ENSURE Parent function $\mathrm{pa} : M \setminus \{r\} \to M$ (spanning tree)
    \FOR{$(f, m) \in M \times M$, $f \neq m$}
    \STATE Compute $s(f \to m)$ using Eq.~\ref{eq:m9-weight}
    \ENDFOR
    \STATE $G \gets$ the directed graph on vertex set $M$ with edge weights $s$, with every edge into $r$ removed
    \STATE $\mathrm{pa} \gets$ the maximum spanning directed tree rooted at $r$ on $G$, found with the Chu--Liu/Edmonds algorithm
    \RETURN $\mathrm{pa}$
  \end{algorithmic}
\end{algorithm}

\section{Root Estimation}\label{sec:appendix-rootest}

TRACE-Set and TRACE-Edge (Sections~\ref{sec:m8}, \ref{sec:m9}) assume
the root $r$ is known. This appendix gives a method for estimating $r$
when it is not, and reports its estimation accuracy and its effect on
tree-recovery performance.

\subsection{Method}\label{sec:appendix-rootest-method}

For the probe example set $P_\tau$ of task $\tau$, we define the mean
of model $m$'s sequence-averaged log-likelihood as
\begin{equation}
  \bar{S}_m(\tau) = \frac{1}{|P_\tau|} \sum_{(x,y) \in P_\tau} S_m(x,y)
\end{equation}
Fine-tuning $p \to m$ on task $\tau_m$ raises $\bar{S}_m(\tau_m)$
above $\bar{S}_p(\tau_m)$, while it may lower the log-likelihood
$\bar{S}_m(\tau')$ for $\tau' \in \mathcal{T} \setminus \{\tau_m\}$
through forgetting. We therefore expect the variance of
$\bar{S}_m(\tau)$ to grow as more rounds of fine-tuning accumulate. We
define model $m$'s across-task variance as
\begin{equation}
  \sigma^2_m = \frac{1}{|\mathcal{T}|}\sum_{\tau \in \mathcal{T}} (\bar{S}_m(\tau) - \mu_m)^2, \quad
  \mu_m = \frac{1}{|\mathcal{T}|}\sum_{\tau \in \mathcal{T}} \bar{S}_m(\tau)
\end{equation}
We estimate the root as the model $\hat{r}$ with the smallest variance
$\sigma^2_m$, and use it as $r$ for TRACE-Set and TRACE-Edge. Because
the level of log-likelihood differs across tasks, however, the
property that the root has the smallest variance is not guaranteed in
general. We verify its validity experimentally in
Appendix~\ref{sec:appendix-rootest-acc}.

\subsection{Estimation Accuracy}\label{sec:appendix-rootest-acc}

The root estimate $\hat{r}$ matched the true root in 22 of 25 trials
($0.88$) for train and 15 of 25 trials ($0.60$) for held-out (chance
level $0.137$). By configuration, the only misses for train were the 3
trials of DEEP. For held-out, the estimate matched in every trial for
the three HET-family configurations, but missed in every trial for
DEEP and HOMO. As stated in Appendix~\ref{sec:appendix-rootest-method},
the property that the root has the smallest across-task variance is
not guaranteed in general, and it did not hold for these two
configurations.

\subsection{Effect on Tree Recovery}\label{sec:appendix-rootest-recovery}

TRACE-Set, TRACE-Edge, and MST all require the root as prior
knowledge, but combining them with the root-estimation method above
lets them operate even when the root is unknown; we denote these as
TRACE-Set+RootEst, TRACE-Edge+RootEst, and MST+RootEst.
Table~\ref{tab:main-rootest} repeats the setup of Table~\ref{tab:main}
(Section~\ref{sec:recover-eval}) with the estimated root $\hat{r}$ in
place of the true root.

\begin{table*}[htbp]
  \caption{Per-tree parent-match accuracy, and overall parent-match
    accuracy (Acc) and depth-match accuracy (DAcc) across all five
    trees, with the estimated root $\hat{r}$
    (Appendix~\ref{sec:appendix-rootest-method}) supplied in place of
    the true root; compare against Table~\ref{tab:main}. The per-tree
    columns are the mean over 5 seeds. The Acc and DAcc columns are the
    mean $\pm$ standard deviation over all 25 runs (5 trees $\times$ 5
    seeds). Bold marks the best value within the same probe.}
  \label{tab:main-rootest}
  \begin{center}
    \small
    \resizebox{\linewidth}{!}{%
      \begin{tabular}{llccccccc}
        \toprule
              &                       & \multicolumn{5}{c}{Per-tree parent-match accuracy} & \multicolumn{2}{c}{All 5 trees}                                                                                                  \\
        \cmidrule(lr){3-7} \cmidrule(lr){8-9}
        probe & Method                & HET-3B                                             & HET-8B                          & HET-QWEN       & DEEP           & HOMO           & Acc                  & DAcc                 \\
        \midrule
        \multirow{4}{*}{train}
              & Star baseline+RootEst & 0.333                                              & 0.333                           & 0.333          & 0.100          & 0.333          & $0.287\pm0.110$      & $0.317\pm0.034$      \\
              & MST+RootEst           & 0.700                                              & 0.667                           & 0.633          & 0.450          & 0.500          & $0.590\pm0.173$      & $0.493\pm0.249$      \\
              & TRACE-Set+RootEst     & \textbf{0.833}                                     & \textbf{1.000}                  & \textbf{0.900} & 0.375          & \textbf{0.800} & $\bm{0.782\pm0.272}$ & $\bm{0.832\pm0.198}$ \\
              & TRACE-Edge+RootEst    & 0.733                                              & 0.667                           & 0.833          & \textbf{0.650} & 0.667          & $0.710\pm0.103$      & $0.672\pm0.139$      \\
        \midrule
        \multirow{4}{*}{held-out}
              & Star baseline+RootEst & 0.333                                              & 0.333                           & 0.333          & 0.000          & 0.333          & $0.267\pm0.136$      & $0.283\pm0.068$      \\
              & MST+RootEst           & 0.633                                              & \textbf{0.567}                  & 0.400          & 0.300          & 0.267          & $0.433\pm0.180$      & $0.368\pm0.175$      \\
              & TRACE-Set+RootEst     & 0.533                                              & 0.333                           & 0.400          & 0.275          & 0.467          & $0.402\pm0.122$      & $\bm{0.453\pm0.228}$ \\
              & TRACE-Edge+RootEst    & \textbf{0.667}                                     & 0.233                           & \textbf{0.600} & \textbf{0.375} & \textbf{0.500} & $\bm{0.475\pm0.199}$ & $0.425\pm0.323$      \\
        \bottomrule
      \end{tabular}}
  \end{center}
\end{table*}
\vspace{-4pt}

\begin{itemize}
  \item \textbf{Under train, estimating the root causes only a small
          drop.} TRACE-Edge's parent-match accuracy falls only from
        $0.740$ to $0.710$, and TRACE-Set's from $0.837$ to $0.782$
        (Table~\ref{tab:main}). The drop is concentrated in DEEP, where
        root estimation missed (TRACE-Set $0.650 \to 0.375$ and
        TRACE-Edge $0.800 \to 0.650$).
  \item \textbf{TRACE-Edge and TRACE-Set still outperform MST with the
          estimated root.} Under train, the difference between
        TRACE-Edge and MST, paired over the 25 trees, is $+0.120$ (SE
        $0.036$). Under held-out, the difference is $+0.042$: smaller
        than under train, but still positive.
\end{itemize}

\section{Experimental Details}\label{sec:tree-detail}

\subsection{Model Trees}\label{sec:appendix-trees}

\begin{figure*}[htbp]
  \centering
  \tikzset{
    mn/.style={draw, rounded corners=1.5pt, inner sep=1.2pt,
        font=\scriptsize, minimum height=3.4mm, minimum width=7mm},
    rt/.style={mn, fill=black!10},
    ed/.style={-latex, semithick},
    el/.style={font=\tiny, midway, above, sloped, inner sep=1pt}
  }
  \begin{minipage}[b]{0.32\linewidth}
    \centering
    \begin{tikzpicture}
      \node[rt] (r)  at (0,-1.25)   {$r$};
      \node[mn] (gm) at (1.50,-0.25) {$g_{\mathrm{math}}$};
      \node[mn] (gs) at (1.50,-1.75) {$g_{\mathrm{sum}}$};
      \node[mn] (c1) at (3.20,0)     {$c_1$};
      \node[mn] (c2) at (3.20,-0.55) {$c_2$};
      \node[mn] (c3) at (3.20,-1.40) {$c_3$};
      \node[mn] (c4) at (3.20,-2.30) {$c_4$};
      \draw[ed] (r)  -- (gm) node[el] {gsm8k};
      \draw[ed] (r)  -- (gs) node[el] {xsum};
      \draw[ed] (gm) -- (c1) node[el] {e2e\_nlg};
      \draw[ed] (gm) -- (c2) node[el] {wmt14};
      \draw[ed] (gs) -- (c3) node[el] {e2e\_nlg};
      \draw[ed] (gs) -- (c4) node[el] {wmt14};
    \end{tikzpicture}\\[2pt]
    {\footnotesize (a) HET-3B ($r$ = Llama-3.2-3B)}
  \end{minipage}
  \hfill
  \begin{minipage}[b]{0.32\linewidth}
    \centering
    \begin{tikzpicture}
      \node[rt] (r)  at (0,-1.25)   {$r$};
      \node[mn] (gm) at (1.50,-0.25) {$g_{\mathrm{math}}$};
      \node[mn] (gs) at (1.50,-1.75) {$g_{\mathrm{sum}}$};
      \node[mn] (c1) at (3.20,0)     {$c_1$};
      \node[mn] (c2) at (3.20,-0.55) {$c_2$};
      \node[mn] (c3) at (3.20,-1.40) {$c_3$};
      \node[mn] (c4) at (3.20,-2.30) {$c_4$};
      \draw[ed] (r)  -- (gm) node[el] {gsm8k};
      \draw[ed] (r)  -- (gs) node[el] {xsum};
      \draw[ed] (gm) -- (c1) node[el] {e2e\_nlg};
      \draw[ed] (gm) -- (c2) node[el] {wmt14};
      \draw[ed] (gs) -- (c3) node[el] {e2e\_nlg};
      \draw[ed] (gs) -- (c4) node[el] {wmt14};
    \end{tikzpicture}\\[2pt]
    {\footnotesize (b) HET-8B ($r$ = Llama-3.1-8B)}
  \end{minipage}
  \hfill
  \begin{minipage}[b]{0.32\linewidth}
    \centering
    \begin{tikzpicture}
      \node[rt] (r)  at (0,-1.25)   {$r$};
      \node[mn] (gm) at (1.50,-0.25) {$g_{\mathrm{math}}$};
      \node[mn] (gs) at (1.50,-1.75) {$g_{\mathrm{sum}}$};
      \node[mn] (c1) at (3.20,0)     {$c_1$};
      \node[mn] (c2) at (3.20,-0.55) {$c_2$};
      \node[mn] (c3) at (3.20,-1.40) {$c_3$};
      \node[mn] (c4) at (3.20,-2.30) {$c_4$};
      \draw[ed] (r)  -- (gm) node[el] {gsm8k};
      \draw[ed] (r)  -- (gs) node[el] {xsum};
      \draw[ed] (gm) -- (c1) node[el] {e2e\_nlg};
      \draw[ed] (gm) -- (c2) node[el] {wmt14};
      \draw[ed] (gs) -- (c3) node[el] {e2e\_nlg};
      \draw[ed] (gs) -- (c4) node[el] {wmt14};
    \end{tikzpicture}\\[2pt]
    {\footnotesize (c) HET-QWEN ($r$ = Qwen2.5-3B)}
  \end{minipage}

  \vspace{8pt}

  \begin{minipage}[b]{0.55\linewidth}
    \centering
    \begin{tikzpicture}
      \node[rt] (r)  at (0,-1.35)   {$r$};
      \node[mn] (gm) at (1.50,-0.25) {$g_{\mathrm{math}}$};
      \node[mn] (gs) at (1.50,-1.95) {$g_{\mathrm{sum}}$};
      \node[mn] (c1) at (3.20,0)     {$c_1$};
      \node[mn] (c2) at (3.20,-0.55) {$c_2$};
      \node[mn] (c3) at (3.20,-1.45) {$c_3$};
      \node[mn] (c4) at (3.20,-2.55) {$c_4$};
      \node[mn] (d1) at (5.00,-1.15) {$d_1$};
      \node[mn] (d2) at (5.00,-1.75) {$d_2$};
      \draw[ed] (r)  -- (gm) node[el] {gsm8k};
      \draw[ed] (r)  -- (gs) node[el] {xsum};
      \draw[ed] (gm) -- (c1) node[el] {e2e\_nlg};
      \draw[ed] (gm) -- (c2) node[el] {wmt14};
      \draw[ed] (gs) -- (c3) node[el] {e2e\_nlg};
      \draw[ed] (gs) -- (c4) node[el] {wmt14};
      \draw[ed] (c3) -- (d1) node[el] {dialogsum};
      \draw[ed] (c3) -- (d2) node[el] {cnn\_dm};
    \end{tikzpicture}\\[2pt]
    {\footnotesize (d) DEEP ($r$ = Llama-3.2-3B, depth 3, 9 nodes)}
  \end{minipage}
  \hfill
  \begin{minipage}[b]{0.42\linewidth}
    \centering
    \begin{tikzpicture}
      \node[rt] (r)  at (0,-0.9)    {$r$};
      \node[mn] (g1) at (1.75,-0.3) {$g_{\mathrm{sum1}}$};
      \node[mn] (g2) at (1.75,-1.5) {$g_{\mathrm{sum2}}$};
      \node[mn] (c1) at (3.70,0)     {$c_1$};
      \node[mn] (c2) at (3.70,-0.6)  {$c_2$};
      \node[mn] (c3) at (3.70,-1.2)  {$c_3$};
      \node[mn] (c4) at (3.70,-1.8)  {$c_4$};
      \draw[ed] (r)  -- (g1) node[el] {xsum};
      \draw[ed] (r)  -- (g2) node[el] {cnn\_dm};
      \draw[ed] (g1) -- (c1) node[el] {dialogsum};
      \draw[ed] (g1) -- (c2) node[el] {samsum};
      \draw[ed] (g2) -- (c3) node[el] {dialogsum};
      \draw[ed] (g2) -- (c4) node[el] {samsum};
    \end{tikzpicture}\\[2pt]
    {\footnotesize (e) HOMO ($r$ = Llama-3.2-3B, all nodes summarization)}
  \end{minipage}

  \caption{Structures and task assignments of the five model trees
    used for evaluation. Shaded nodes are the root $r$, and edge
    labels denote the task $\tau_m$ the child node used for
    fine-tuning. (a)--(c) share the same structure and task
    assignment, differing only in the pretrained model placed at the
    root. (d) derives two depth-3 nodes $d_1, d_2$ from $c_3$ of (a),
    and (e) fine-tunes every node on summarization tasks only. wmt14
    abbreviates wmt14\_de\_en, and cnn\_dm abbreviates
    cnn\_dailymail.}
  \label{fig:trees-all}
\end{figure*}
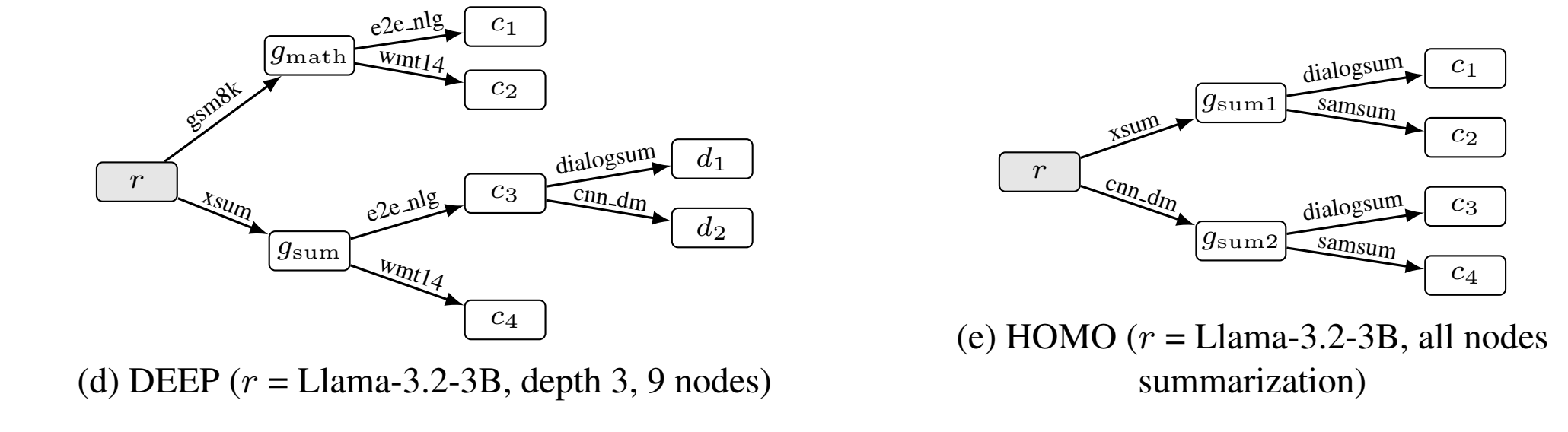

Table~\ref{tab:trees} lists the five model trees used for evaluation.
All share the same basic form: a 7-node tree consisting of the root, 2
nodes at depth 1, and 4 nodes at depth 2, with DEEP alone adding 2
nodes at depth 3 for 9 nodes in total. At each level, the 2 nodes of
that level each serve as the parent of children fine-tuned on the 2
tasks listed in the table. Every internal node other than the root
therefore has exactly 2 children. Figure~\ref{fig:trees-all} shows the
structure and task assignment of all five trees.

\begin{table*}[htbp]
  \caption{The five model trees used for evaluation. The ``Depth $k$''
    columns list the tasks each node at depth $k$ used for
    fine-tuning. At depths 1 and 2, each of the 2 nodes one level up
    derives children fine-tuned on the 2 listed tasks. The 2 depth-3
    nodes of DEEP are children of $c_3$, the depth-2 e2e\_nlg child.}
  \label{tab:trees}
  \begin{center}
    \resizebox{\linewidth}{!}{%
      \begin{tabular}{llccllll}
        \toprule
        Tree     & Root model                  & Nodes & Depth & Depth 1              & Depth 2                 & Depth 3                   & Change from reference tree \\
        \midrule
        HET-3B   & Llama-3.2-3B~\citep{llama3} & 7     & 2     & gsm8k, xsum          & e2e\_nlg, wmt14\_de\_en & ---                       & --- (reference tree)       \\
        HET-8B   & Llama-3.1-8B~\citep{llama3} & 7     & 2     & gsm8k, xsum          & e2e\_nlg, wmt14\_de\_en & ---                       & Root model size            \\
        HET-QWEN & Qwen2.5-3B~\citep{qwen25}   & 7     & 2     & gsm8k, xsum          & e2e\_nlg, wmt14\_de\_en & ---                       & Root model family          \\
        DEEP     & Llama-3.2-3B~\citep{llama3} & 9     & 3     & gsm8k, xsum          & e2e\_nlg, wmt14\_de\_en & dialogsum, cnn\_dailymail & Tree depth                 \\
        HOMO     & Llama-3.2-3B~\citep{llama3} & 7     & 2     & xsum, cnn\_dailymail & dialogsum, samsum       & ---                       & Diversity of child tasks   \\
        \bottomrule
      \end{tabular}}
  \end{center}
\end{table*}
\vspace{-4pt}

\vspace{-4pt}

\subsection{Models and Datasets Used}\label{sec:appendix-data}

We use three pretrained models as roots: Llama-3.2-3B, Llama-3.1-8B~\citep{llama3},
and Qwen2.5-3B~\citep{qwen25}, all base models without instruction
tuning. Table~\ref{tab:datasets} lists the datasets used for
fine-tuning and for measuring the deviation. We use each dataset's
training split for fine-tuning, and construct the probe set
(Section~\ref{sec:tree-setup}) by drawing 100 examples per task from
the same training split.

\begin{table*}[htbp]
  \caption{Datasets used for fine-tuning and for the train/held-out
    probe sets.}
  \label{tab:datasets}
  \begin{center}
    \small
    \begin{tabular}{ll}
      \toprule
      Dataset                      & Task type                        \\
      \midrule
      gsm8k~\citep{gsm8k}          & Math word problems               \\
      xsum~\citep{xsum}            & Summarization                    \\
      e2e\_nlg~\citep{e2enlg}      & Data-to-text generation          \\
      wmt14\_de\_en~\citep{wmt14}  & Translation (German$\to$English) \\
      dialogsum~\citep{dialogsum}  & Dialogue summarization           \\
      cnn\_dailymail~\citep{cnndm} & Summarization                    \\
      samsum~\citep{samsum}        & Dialogue summarization           \\
      \bottomrule
    \end{tabular}
  \end{center}
\end{table*}
\vspace{-4pt}

\subsection{Training Configuration}\label{sec:appendix-hp}

We fine-tune every node with LoRA~\citep{lora}, using hyperparameters
shared across all tasks and trees (learning rate $2\times10^{-4}$, 5
epochs, LoRA rank $=8$, $\alpha=16$, targeting q\_proj and v\_proj). We
confirmed the validity of sharing these hyperparameters beforehand by
comparing against a per-task search. The accuracy gap against
task-specific settings was at most 0.025 (gsm8k), and at most 0.02 for
the other tasks. We constructed five trees per configuration, with
random seeds ranging from 0 to 4. Every metric in this paper is
reported as the mean and standard deviation over these 5 seeds.

We choose TRACE-Set's threshold $\eta$ in Eq.~\ref{eq:ancestor-set}
independently for each configuration from the candidate set
$\{0, 0.05, 0.1, 0.2\}$ by leave-one-tree-out (selecting the value that
maximizes the mean parent-match accuracy over the remaining 4 trees and
5 seeds, excluding the tree under evaluation). For train, we used
$\eta=0.1$ for every configuration. For held-out, we used $0.05$ for
HET-3B, $0.2$ for DEEP, and $0.1$ for the remaining 3 configurations.

\subsection{Verification on a Full-FT Tree}\label{sec:appendix-fullft}

The main results (Section~\ref{sec:recover-eval}) cover the five trees
fine-tuned with LoRA. To check that limiting fine-tuning to LoRA does
not bias the results, we additionally evaluated one tree fine-tuned
with all parameters updated (Full FT). Its topology is identical to
HET-3B (root Llama-3.2-3B, 7 nodes), and we updated all parameters
with a learning rate of $2\times10^{-5}$. We used the same value of
TRACE-Set's threshold $\eta$ chosen for the five LoRA trees
($\eta=0.1$) without change.

Table~\ref{tab:fullft} shows the results. Under train, both TRACE-Set
and TRACE-Edge reach a parent-match accuracy of $1.000$, the same
trend as the five LoRA trees (Table~\ref{tab:main}). Under held-out,
MST ($0.867$) gives the best result. Because this is a single
tree, we treat it only as a check on the trend and do not include it
in the main results.

\begin{table*}[htbp]
  \centering
  \caption{Parent-match accuracy (Acc) and depth-match accuracy (DAcc)
    on the Full FT tree (single tree, 5 seeds), mean $\pm$ SD. Bold
    marks the best value per column.}
  \label{tab:fullft}
  \small
  \begin{tabular}{lcccc}
    \toprule
    Method          & train Acc            & train DAcc           & held-out Acc         & held-out DAcc        \\
    \midrule
    Star baseline   & $0.333\pm0.000$      & $0.333\pm0.000$      & $0.333\pm0.000$      & $0.333\pm0.000$      \\
    MST (black-box) & $0.967\pm0.075$      & $0.967\pm0.075$      & $\bm{0.867\pm0.183}$ & $\bm{0.800\pm0.274}$ \\
    TRACE-Set       & $\bm{1.000\pm0.000}$ & $\bm{1.000\pm0.000}$ & $0.333\pm0.000$      & $0.500\pm0.000$      \\
    TRACE-Edge      & $\bm{1.000\pm0.000}$ & $\bm{1.000\pm0.000}$ & $0.500\pm0.000$      & $0.500\pm0.000$      \\
    \bottomrule
  \end{tabular}
  \vspace{1mm}
\end{table*}
\vspace{-4pt}

\section{Details under Different Probe Conditions}\label{sec:appendix-probe}

This section details the probe conditions discussed in
Section~\ref{sec:limitation}. train uses the training examples each
node used for fine-tuning, and held-out uses non-training examples
drawn from the same distribution as each task's training examples.

To examine the properties of the deviation under each probe condition,
we define the deviation matrix. $\delta_\tau(f \to m)$ is the change
in fit to task $\tau$ of $m$ relative to model $f$, and when
$\delta_\tau(f \to m) > 0$, we consider it likely that $\tau$ was used
to fine-tune $m$ somewhere along its derivation from $f$. Arranging
this for every node and task,
\begin{equation}
  \Delta(f) =
  \begin{pmatrix}
    \delta_{\tau_1}(f \to m_1) & \delta_{\tau_2}(f \to m_1) & \cdots & \delta_{\tau_T}(f \to m_1) \\
    \delta_{\tau_1}(f \to m_2) & \delta_{\tau_2}(f \to m_2) & \cdots & \delta_{\tau_T}(f \to m_2) \\
    \vdots                     & \vdots                     & \ddots & \vdots                     \\
    \delta_{\tau_1}(f \to m_N) & \delta_{\tau_2}(f \to m_N) & \cdots & \delta_{\tau_T}(f \to m_N)
  \end{pmatrix}
  \label{eq:delta-matrix}
\end{equation}
we call this the deviation matrix relative to $f$, where $M \setminus
  \{f\} = \{m_1,\ldots,m_N\}$.

\begin{figure*}[htbp]
  \centering
  \includegraphics[width=\linewidth]{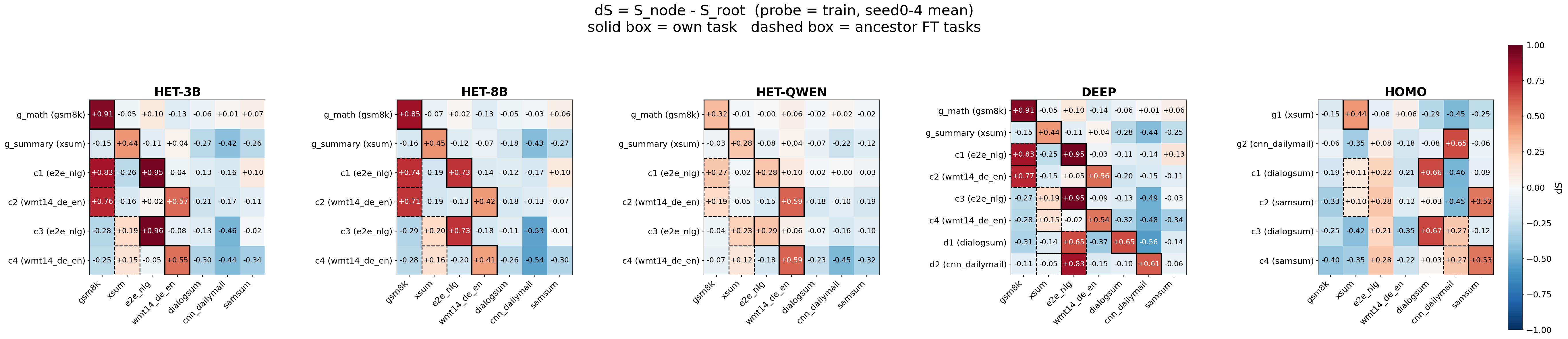}
  \caption{Node-by-task deviation matrices for the five model trees
    (train probe set).}
  \label{fig:heatmap-train}
\end{figure*}

\begin{figure*}[htbp]
  \centering
  \includegraphics[width=\linewidth]{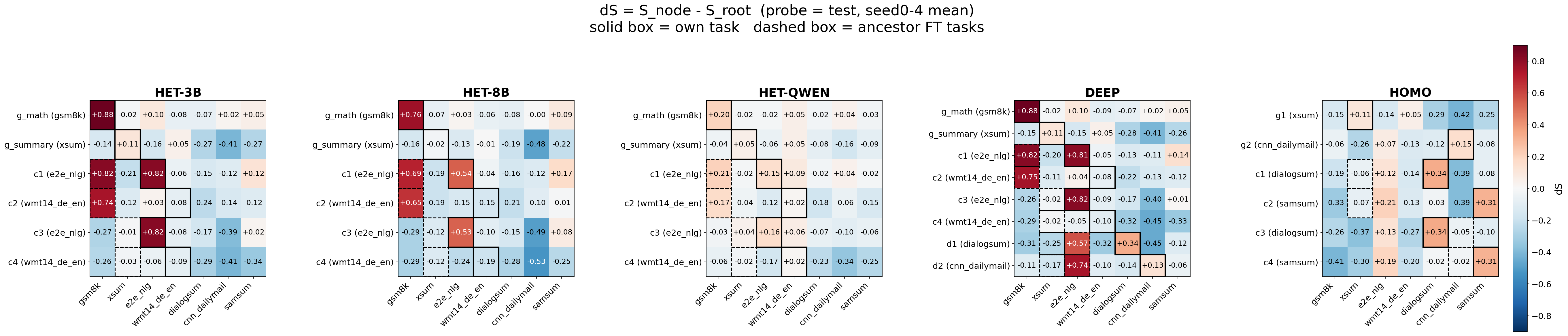}
  \caption{Deviation matrices with the probe set set to held-out
    (non-training examples from the same distribution). The positive
    deviation for a node's own fine-tuning task remains for most
    nodes, but the deviation for ancestor tasks fails to stay
    positive.}
  \label{fig:heatmap-held-out}
\end{figure*}

Figures~\ref{fig:heatmap-train} and \ref{fig:heatmap-held-out} show
the deviation matrices for train and held-out.

\vspace{-4pt}

\end{document}

%% file: math_commands.tex
\usepackage{amsmath,amsfonts,bm}

\def\eqref#1{equation~\ref{#1}}

\def\1{\bm{1}}

\DeclareMathAlphabet{\mathsfit}{\encodingdefault}{\sfdefault}{m}{sl}
\SetMathAlphabet{\mathsfit}{bold}{\encodingdefault}{\sfdefault}{bx}{n}

\DeclareMathOperator*{\argmax}{arg\,max}

%% file: main.bbl
\begin{thebibliography}{22}
\providecommand{\natexlab}[1]{#1}
\providecommand{\url}[1]{\texttt{#1}}
\expandafter\ifx\csname urlstyle\endcsname\relax
  \providecommand{\doi}[1]{doi: #1}\else
  \providecommand{\doi}{doi: \begingroup \urlstyle{rm}\Url}\fi

\bibitem[Bojar et~al.(2014)Bojar, Buck, Federmann, et~al.]{wmt14}
Ond{\v{r}}ej Bojar, Christian Buck, Christian Federmann, et~al.
\newblock Findings of the 2014 workshop on statistical machine translation.
\newblock In \emph{Proceedings of the Ninth Workshop on Statistical Machine
  Translation (WMT)}, pp.\  12--58, 2014.

\bibitem[Chen et~al.(2021)Chen, Liu, Chen, et~al.]{dialogsum}
Yulong Chen, Yang Liu, Liang Chen, et~al.
\newblock Dialog{S}um: A real-life scenario dialogue summarization dataset.
\newblock In \emph{Findings of the Association for Computational Linguistics:
  ACL-IJCNLP 2021}, pp.\  5062--5074, 2021.

\bibitem[Chu \& Liu(1965)Chu and Liu]{chu1965shortest}
Yoeng-Jin Chu and Tseng-Hong Liu.
\newblock On the shortest arborescence of a directed graph.
\newblock \emph{Scientia Sinica}, 14\penalty0 (10):\penalty0 1396--1400, 1965.

\bibitem[Cobbe et~al.(2021)Cobbe, Kosaraju, Bavarian, et~al.]{gsm8k}
Karl Cobbe, Vineet Kosaraju, Mohammad Bavarian, et~al.
\newblock Training verifiers to solve math word problems, 2021.
\newblock arXiv preprint arXiv:2110.14168.

\bibitem[Edmonds(1967)]{edmonds1967optimum}
Jack Edmonds.
\newblock Optimum branchings.
\newblock \emph{Journal of Research of the National Bureau of Standards,
  Section B: Mathematics and Mathematical Physics}, 71B\penalty0 (4):\penalty0
  233--240, 1967.
\newblock \doi{10.6028/jres.071B.032}.

\bibitem[Gliwa et~al.(2019)Gliwa, Mochol, Biesek, et~al.]{samsum}
Bogdan Gliwa, Iwona Mochol, Maciej Biesek, et~al.
\newblock {SAMS}um corpus: A human-annotated dialogue dataset for abstractive
  summarization.
\newblock In \emph{Proceedings of the 2nd Workshop on New Frontiers in
  Summarization}, pp.\  70--79, 2019.

\bibitem[Grattafiori et~al.(2024)Grattafiori, Dubey, Jauhri, et~al.]{llama3}
Aaron Grattafiori, Abhimanyu Dubey, Abhinav Jauhri, et~al.
\newblock The llama 3 herd of models, 2024.
\newblock arXiv preprint arXiv:2407.21783.

\bibitem[Hermann et~al.(2015)Hermann, Ko{\v{c}}isk{\'y}, Grefenstette,
  et~al.]{cnndm}
Karl~Moritz Hermann, Tom{\'a}{\v{s}} Ko{\v{c}}isk{\'y}, Edward Grefenstette,
  et~al.
\newblock Teaching machines to read and comprehend.
\newblock In \emph{Advances in Neural Information Processing Systems (NIPS)},
  2015.

\bibitem[Horwitz et~al.(2025)Horwitz, Shul, and
  Hoshen]{horwitz2025unsupervised}
Eliahu Horwitz, Asaf Shul, and Yedid Hoshen.
\newblock Unsupervised model tree heritage recovery.
\newblock In \emph{International Conference on Learning Representations},
  volume 2025, 2025.

\bibitem[Hu et~al.(2022)Hu, Shen, Wallis, et~al.]{lora}
Edward~J. Hu, Yelong Shen, Phillip Wallis, et~al.
\newblock Lo{RA}: Low-rank adaptation of large language models.
\newblock In \emph{International Conference on Learning Representations
  (ICLR)}, 2022.

\bibitem[{Hugging Face}(2023)]{huggingface_tgi}
{Hugging Face}.
\newblock Text generation inference.
\newblock \url{https://github.com/huggingface/text-generation-inference}, 2023.
\newblock Accessed: 2026-09-26.

\bibitem[Kwon et~al.(2023)Kwon, Li, Zhuang, Sheng, Zheng, Yu, Gonzalez, Zhang,
  and Stoica]{kwon2023efficient}
Woosuk Kwon, Zhuohan Li, Siyuan Zhuang, Ying Sheng, Lianmin Zheng, Cody~Hao Yu,
  Joseph Gonzalez, Hao Zhang, and Ion Stoica.
\newblock Efficient memory management for large language model serving with
  pagedattention.
\newblock In \emph{Proceedings of the 29th symposium on operating systems
  principles}, pp.\  611--626, 2023.

\bibitem[Maini et~al.(2024)Maini, Jia, Papernot, and Dziedzic]{maini2024llm}
Pratyush Maini, Hengrui Jia, Nicolas Papernot, and Adam Dziedzic.
\newblock {LLM} dataset inference: Did you train on my dataset?
\newblock In \emph{The Thirty-eighth Annual Conference on Neural Information
  Processing Systems}, 2024.
\newblock arXiv:2406.06443.

\bibitem[Narayan et~al.(2018)Narayan, Cohen, and Lapata]{xsum}
Shashi Narayan, Shay~B. Cohen, and Mirella Lapata.
\newblock Don't give me the details, just the summary! topic-aware
  convolutional neural networks for extreme summarization.
\newblock In \emph{Proceedings of the 2018 Conference on Empirical Methods in
  Natural Language Processing (EMNLP)}, pp.\  1797--1807, 2018.

\bibitem[Nikolic et~al.(2025)Nikolic, Baluta, and Saxena]{nikolic2025model}
Ivica Nikolic, Teodora Baluta, and Prateek Saxena.
\newblock Model provenance testing for large language models.
\newblock In \emph{The Thirty-ninth Annual Conference on Neural Information
  Processing Systems}, 2025.
\newblock URL \url{https://openreview.net/forum?id=Iy4cAXotrf}.

\bibitem[Novikova et~al.(2017)Novikova, Du{\v{s}}ek, and Rieser]{e2enlg}
Jekaterina Novikova, Ond{\v{r}}ej Du{\v{s}}ek, and Verena Rieser.
\newblock The e2e dataset: New challenges for end-to-end generation.
\newblock In \emph{Proceedings of the 18th Annual SIGdial Meeting on Discourse
  and Dialogue}, pp.\  201--206, 2017.

\bibitem[Peng et~al.(2023)Peng, Yi, Wu, Wu, Zhu, Lyu, Jiao, Xu, Sun, and
  Xie]{peng2023you}
Wenjun Peng, Jingwei Yi, Fangzhao Wu, Shangxi Wu, Bin~Bin Zhu, Lingjuan Lyu,
  Binxing Jiao, Tong Xu, Guangzhong Sun, and Xing Xie.
\newblock Are you copying my model? protecting the copyright of large language
  models for eaas via backdoor watermark.
\newblock In \emph{Proceedings of the 61st Annual Meeting of the Association
  for Computational Linguistics (Volume 1: Long Papers)}, pp.\  7653--7668,
  2023.

\bibitem[Tan et~al.(2025)Tan, Yu, Wen, and
  Sakuma]{tan2025leakawaypretrainedmodel}
Yixin Tan, Zhe Yu, Rui Wen, and Jun Sakuma.
\newblock One leak away: How pretrained model exposure amplifies jailbreak
  risks in finetuned llms.
\newblock \emph{arXiv preprint arXiv:2512.14751}, 2025.

\bibitem[Yang et~al.(2024)Yang, Yang, Zhang, et~al.]{qwen25}
An~Yang, Baosong Yang, Beichen Zhang, et~al.
\newblock Qwen2.5 technical report, 2024.
\newblock arXiv preprint arXiv:2412.15115.

\bibitem[Yax et~al.(2025)Yax, Oudeyer, and Palminteri]{phylolm}
Nicolas Yax, Pierre-Yves Oudeyer, and Stefano Palminteri.
\newblock Phylolm: Inferring the phylogeny of large language models and
  predicting their performances in benchmarks.
\newblock In \emph{International Conference on Learning Representations
  (ICLR)}, 2025.

\bibitem[Zeng et~al.(2024)Zeng, Wang, Hu, Xu, Zhou, Wang, Yu, and
  Lin]{zeng2024huref}
Boyi Zeng, Lizheng Wang, Yuncong Hu, Yi~Xu, Chenghu Zhou, Xinbing Wang, Yu~Yu,
  and Zhouhan Lin.
\newblock Huref: {HU}man-{RE}adable fingerprint for large language models.
\newblock In \emph{The Thirty-eighth Annual Conference on Neural Information
  Processing Systems}, 2024.
\newblock URL \url{https://openreview.net/forum?id=ibggY9ZJ1T}.

\bibitem[Zhong \& Raghunathan(2026)Zhong and Raghunathan]{zhong2026watch}
Ziqian Zhong and Aditi Raghunathan.
\newblock Watch the weights: Unsupervised monitoring and control of fine-tuned
  llms.
\newblock In \emph{International Conference on Learning Representations},
  volume 2026, pp.\  75198--75234, 2026.

\end{thebibliography}
